\documentclass[10pt,twocolumn,letterpaper]{article}

\usepackage{cvpr}              % To produce the CAMERA-READY version
\usepackage[dvipsnames]{xcolor}

\usepackage[ruled,linesnumbered]{algorithm2e}
\usepackage{multirow}
\usepackage{bbding}
\usepackage{diagbox}
\usepackage{ulem}
\usepackage{colortbl}
\usepackage{makecell}
\definecolor{cvprblue}{rgb}{0.21,0.49,0.74}
\usepackage[pagebackref,breaklinks,colorlinks,citecolor=cvprblue]{hyperref}

\def\paperID{13502} % *** Enter the Paper ID here
\def\confName{CVPR}
\def\confYear{2024}

\title{Towards Accurate and Robust Architectures via Neural Architecture Search}

\makeatletter
\newcommand{\printfnsymbol}[1]{%
	\textsuperscript{\@fnsymbol{#1}}%
}
\makeatother

\author{Yuwei Ou\thanks{Equal contribution.}, Yuqi Feng\printfnsymbol{1}, Yanan Sun\thanks{Corresponding author.} \\
	College of Computer Science, Sichuan University. \\
	\small\texttt{ouyuwei@stu.scu.edu.cn; feng770623@gmail.com; ysun@scu.edu.cn}
}

\begin{document}
\maketitle
\begin{abstract}
To defend deep neural networks from adversarial attacks, adversarial training has been drawing increasing attention for its effectiveness. However, the accuracy and robustness resulting from the adversarial training are limited by the architecture, because adversarial training improves accuracy and robustness by adjusting the weight connection affiliated to the architecture. In this work, we propose ARNAS to search for accurate and robust architectures for adversarial training. First we design an accurate and robust search space, in which the placement of the cells and the proportional relationship of the filter numbers are carefully determined. With the design, the architectures can obtain both accuracy and robustness by deploying accurate and robust structures to their sensitive positions, respectively. Then we propose a differentiable multi-objective search strategy, performing gradient descent towards directions that are beneficial for both natural loss and adversarial loss, thus the accuracy and robustness can be guaranteed at the same time. We conduct comprehensive experiments in terms of white-box attacks, black-box attacks, and transferability. Experimental results show that the searched architecture has the strongest robustness with the competitive accuracy, and breaks the traditional idea that NAS-based architectures cannot transfer well to complex tasks in robustness scenarios. By analyzing outstanding architectures searched, we also conclude that accurate and robust neural architectures tend to deploy different structures near the input and output, which has great practical significance on both hand-crafting and automatically designing of accurate and robust architectures.

\end{abstract}    
\section{Introduction}
\label{sec:intro}
\begin{figure}
	\includegraphics[width=\columnwidth]{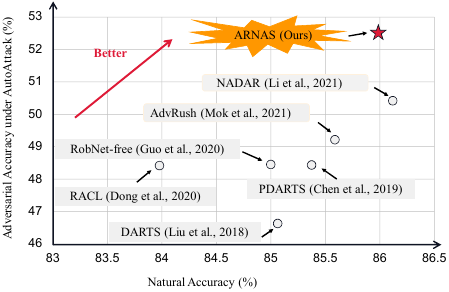}
	\caption{Natural and adversarial accuracy on CIFAR-10. All the architectures are adversarially trained using 7-step PGD, and the adversarial accuracy is evaluated under AutoAttack.}\label{compare}
\end{figure}
Deep neural networks (DNNs) have shown remarkable performance in various real-world applications~\cite{krizhevsky2012imagenet,huang2017densely,bahdanau2014neural}. However, DNNs are found to be vulnerable to adversarial attacks~\cite{szegedy2013intriguing}. The phenomenon limits the application of DNNs in security-critical systems such as self-driving cars~\cite{eykholt2018robust} and face recognition~\cite{sharif2016accessorize}. After the discovery of this intriguing weakness of DNNs, various methods have been proposed to defend DNNs from adversarial attacks. Among them, adversarial training~\cite{madry2017towards,goodfellow2014explaining} has been widely used for its excellent ability to enhance robustness~\cite{zhang2019theoretically,mok2021advrush}. Different from the standard training that trains with natural data directly, the adversarial training trains the network with adversarial examples. However, the problem is that the accuracy and robustness resulting from the adversarial training are limited if the architecture is not well designed in advance~\cite{guo2020meets}, because the adversarial training improves the accuracy and robustness by adjusting the weight connection that is affiliated to the architecture.

Recently, an advanced research topic, known as robust NAS~\cite{guo2020meets,dong2020adversarially,mok2021advrush}, has been widely investigated to solve this problem. Specifically, robust NAS designs intrinsically accurate and robust neural architectures utilizing neural architecture search (NAS)~\cite{zoph2016neural}. These architectures have been validated to show strong accuracy and robustness after adversarial training. For example, RobNet~\cite{guo2020meets} adopts the one-shot NAS~\cite{bender2018understanding} method, adversarially trains a super-network for once and then finds out the robust sub-networks by evaluating each of them under adversarial attacks using the shared weights. ABanditNAS~\cite{chen2020anti} introduces an anti-bandit algorithm, and searches for robust architectures by gradually abandoning operations that are not likely to make the architectures robust. NADAR~\cite{li2021neural} proposes an architecture dilation method, begins with the backbone network of a satisfactory accuracy over the natural data, and searches for a dilation architecture to pursue a maximal robustness gain while preserving a minimal accuracy drop. Besides, the most common way is to employ the differentiable search method, and search for robust architectures by updating the architectures utilizing some robustness metrics. These robustness metrics include Lipschitz constant~\cite{cisse2017parseval} adopted by RACL~\cite{dong2020adversarially}, input loss landscape~\cite{zhao2020bridging} employed by AdvRush~\cite{mok2021advrush}, certified lower bound and Jacobian norm bound proposed and used by DSRNA~\cite{hosseini2021dsrna}. 

After a comprehensive investigation of existing robust NAS methods, we find that there are still two shortcomings in the existing works. First, most of the existing robust NAS just simply adopt the search space of conventional NAS algorithms designed for standard training~\cite{zoph2018learning}. However, it is explored that the architectures suitable for the adversarial training may have different overall structures from those suitable for the standard training~\cite{huang2021exploring}. Consequently, the search space adopted by existing robust NAS is no longer suitable for the adversarial training. Second, there is a trade-off between accuracy and robustness~\cite{tsipras2018robustness,zhang2019theoretically}. Most of the existing robust NAS solve this multi-objective optimization problem by transforming it to a single-objective optimization problem with a fixed regularization coefficient, and optimizing the problem using gradient descent. However, the optimization results heavily rely on the selection of the coefficients. Meanwhile, scholars studying multi-objective optimization have shown that always finding a descent direction common to all criteria may be better for identifying the Pareto front~\cite{desideri2012multiple,desideri2012mutiple}, while the fixed regularization coefficient cannot realize this.

In this paper, we consider the above two problems comprehensively, and propose the Accurate and Robust Neural Architecture Search (ARNAS) method. As shown in~\cref{compare}, compared with peer competitors, the proposed method significantly improves the robustness while achieving similar accuracy to the best after an identical process of adversarial training. Our contributions can be summarized as follows:
	
	\begin{itemize}
\item We design a novel accurate and robust search space to solve the problem that the conventional search space does not contain accurate and robust architectures for adversarial training. Specifically, motivated by preliminary study~\cite{huang2021exploring} that depth and width in different positions of architectures have different effects on accuracy and robustness, we further conjecture that the architectures themselves in different positions also play different roles. We conduct experiments to support the conjecture, and accordingly design a novel cell-based search space. the designed search space is composed of Accurate Cell, Robust Cell, and Reduction Cell. We determine the placement of the cells and the proportional relationship of the filter numbers through experiments. 
\item We propose a differentiable multi-objective search strategy to address the problem that previous search strategies cannot effectively achieve the dual benefits of accuracy and robustness. Specifically, based on multiple gradient descent method (MGDA), we further design a multi-objective adversarial training method, which first finds a common descent direction of natural loss and adversarial loss by determining their coefficients dynamically and automatically, and then performs gradient descent to optimize the architectures towards smaller natural loss and adversarial loss.
\item We conduct comprehensive experiments in terms of white-box attacks, black-box attacks and transferability. The experiments of white-box and black-box attacks show the strongest robustness and high accuracy of the searched architecture. The experiments of transferability break the prejudice that NAS-based architectures cannot transfer well as the task complexity increases. By analyzing outstanding architectures searched, we also conclude that the architectures can obtain both accuracy and robustness by deploying very different structures in different positions, which has great guiding significance on both hand-crafting and automatically designing of accurate and robust architectures.
\end{itemize}

\section{Related Works}\label{Related Works}
\subsection{Adversarial Attacks and Defenses}
According to whether the attacker has full access to the target model or not, existing adversarial attack methods can be divided into white-box attacks and black-box attacks. In relevant fields, commonly used white-box attacks include FGSM~\cite{goodfellow2014explaining}, C\&W~\cite{carlini2017towards}, and PGD~\cite{madry2017towards}. Commonly used black-box attack is the transfer-based attack~\cite{papernot2017practical}. Recently, AutoAttack~\cite{croce2020reliable}, an ensemble of attacks containing both white-box and black-box attacks, becomes popular for robustness evaluation because it is parameter-free, computationally affordable, and user-independent. We also include AutoAttack in our experiments for reliable comparison.

To defend neural networks from adversarial attacks, numerous defense methods have been proposed, among which the adversarial training \cite{madry2017towards,goodfellow2014explaining} has been the most popular way so far to help neural networks enhance their robustness~\cite{guo2020meets,dong2020adversarially,mok2021advrush,chen2020anti}. To perform the adversarial training, the most widely used method is to replace input data with adversarial examples generated by PGD, for the reason that neural networks adversarially trained using PGD usually generalize to other attacks~\cite{madry2017towards}. Besides, some other methods such as defensive distillation~\cite{papernot2016distillation}, data compression~\cite{dziugaite2016study,guo2017countering}, feature denoising~\cite{xie2019feature}, and model ensemble~\cite{tramer2017ensemble,pang2019improving} also demonstrate their feasibility. In our work, we notice the fact that the adversarial training, as the most effective defense method, depends on the design of neural architectures. If the neural architectures are not suitable, they can only obtain low accuracy and low robustness. Our work tackles the problem by automatically designing neural architectures that perform well after adversarial training.

\subsection{Neural Architecture Search (NAS)}
NAS is a promising technique which aims to automate the architecture design of DNNs. According to the search strategy adopted, NAS can be divided into three categories: evolutionary computation-based~\cite{real2019regularized,real2017large} NAS, reinforcement learning-based~\cite{zoph2016neural} NAS, and the differentiable NAS~\cite{liu2018darts,chen2019progressive,xu2019pc}. Among them, the differentiable NAS is especially popular for designing robust architectures because of its efficiency and effectiveness. In the differentiable NAS methods, the search space is relaxed to be continuous, so that the architectures can be designed by optimizing differentiable metrics using gradient descent. By introducing differentiable metrics of robustness, robust architectures can be searched. Our work in this paper also utilizes the differentiable NAS for its efficiency and effectiveness.
\section{The Proposed ARNAS Method}\label{The Proposed DSARA Method}

\subsection{ARNAS Overview}
\begin{algorithm}
	\SetAlgoLined
	\caption{The Framework of ARNAS Method}\label{Algorithm_overview}
	\KwIn{$E$ $\leftarrow$ Total number of epochs for search}
	\KwOut{$f^{*}_{A}$ $\leftarrow$ Final architecture}
	Construct the \textbf{accurate and robust search space}\\
	
	$f_{super}^{0}$ $\leftarrow$Initialize a supernet based on DARTS according to the constructed search space\\
	\For{$i\leftarrow 1$ \KwTo $E$}{
		$f_{super}^{i}$ $\leftarrow$ Optimize $f_{super}^{i-1}$ using \textbf{the differentiable multi-objective search strategy}
	}
	$f^{*}_{A}$ $\leftarrow$ Apply the discretization rule of DARTS to $f_{super}^{E}$
\end{algorithm}
The framework of the proposed ARNAS method is presented in~\cref{Algorithm_overview}. First, we construct the accurate and robust search space, and then initialize a supernet based on the constructed search space. After that, we iteratively optimize the supernet using the proposed differentiable multi-objective search strategy. Finally, the best architecture can be obtained using the discretization rule of DARTS~\cite{liu2018darts}. The innovations of the proposed ARNAS method lies in the accurate and robust search space and the differentiable multi-objective search strategy, which will be introduced in detail in the following subsections.

\subsection{Accurate and Robust Search Space}\label{Search Space}
In this subsection, we will introduce 1) the characteristics of accurate and robust architectures, 2) the limitation of the conventional search space, and 3) the construction of the accurate and robust search space in turn.  
\subsubsection{The Characteristics of Accurate and Robust Architectures}

In order to design an accurate and robust search space for adversarial training, we should first consider the characteristics of accurate and robust architectures. However, the research of NAS and robust neural networks are emerging topics, there is rare knowledge can be directly explored. We are aware of a recent work~\cite{huang2021exploring} which explores the architectural ingredients of adversarially robust deep neural networks. It experimentally concludes that the depth and the width of the neural architectures in different positions have dissimilar effects on the accuracy and the robustness. Motivated by this, we further make the conjecture shown in Proposition~\ref{as}. The proposition would also be experimentally verified in~\cref{Architectural Ingredients}.

\newtheorem{Proposition}{Proposition}
\begin{Proposition}\label{as}
	The cells in different positions of the overall architecture may have different effects on the accuracy and the robustness, and the accuracy and the robustness of the neural architectures can be improved simultaneously by placing different cells in different position.
\end{Proposition}
\subsubsection{The Limitation of Conventional Search Space}
Based on Proposition~\ref{as}, we further analyze the limitation of the conventional search space. In particular, the conventional cell-based search space refers to the one popularized by the famous DARTS algorithm. Currently, almost all differentiable robust NAS methods adopt this search space~\cite{hosseini2021dsrna,dong2020adversarially,mok2021advrush}. Specifically, the cell-based search space only designs two kinds of computation cells named Normal Cells and Reduction Cells, which play the role of enhancing accuracy and reducing data dimension for improving efficiency, respectively. Based on the design, the overall architecture is constructed by stacking multiple Normal Cells to increase the accuracy as much as possible, and rare Reduction Cells to avoid the invalid data dimension. The resulting architecture is mainly composed of the same Normal Cells, and the architectures matching Proposition~\ref{as} (such as the architectures that employ the separable convolutions near the input but employ the dilated convolutions near the output) are not contained in the search space, which limits the improvement of accuracy and robustness.

\subsubsection{The Construction of the Accurate and Robust Search Space}
Given the above analysis, we retain the Reduction Cell while replacing the single type of Normal Cell with Accurate Cell and Robust Cell, aiming to include the architectures matching Proposition~\ref{as}. Consequently, there are three types of cells designed in the proposed search space: Accurate Cell, Robust Cell, and Reduction Cell. In particular, Accurate Cells and Robust Cells both return a feature map of the same dimension but can be placed in different positions to take the corresponding effects of accurate and robustness, as concluded in Proposition~\ref{as}. Reduction Cells return a feature map where the feature map height and width is reduced by a factor of two, playing the same role as common. With this design, the learned cells could also be stacked to form a full network, resulting the scalability of the designed search space. In the following, we will introduce the details about how to determine their particular position to achieve both accuracy and robustness.

\begin{figure}
	\centerline{\includegraphics[width=\columnwidth]{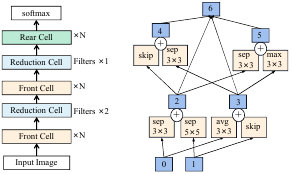}}
	\caption{An example of the proposed search space for CIFAR-10. LEFT: the full outer structure. RIGHT: cell example.}
	\label{fig_search_space}
\end{figure}

As shown in~\cref{fig_search_space}, the Reduction Cells are placed at one-third and two-thirds of the overall architecture. In the rest of the architecture, the Accurate Cells are placed before the second Reduction Cell, while the Robust Cells are placed after the second Reduction Cell. Meanwhile, instead of the common heuristic~\cite{zoph2018learning} that doubling the number of filters in the output whenever the spatial activation size is reduced, we double the number of filters at the first Reduction Cell, and keep the number of filters unchanged at the second Reduction Cell. The effectiveness of this design will be verified in~\cref{Ablation Study}, and the naming rules (i.e., how to determine which cells are for accuracy and which cells are for robustness) are explained in~\cref{Architectural Ingredients}.

\subsection{Differentiable Multi-Objective Search Strategy}\label{Two-Stage Search Strategy}

To search for both accurate and robust architectures from the designed search space, the search process can be formulated by~\cref{eq6}.
\begin{equation}\label{eq6}
	\begin{cases}
		\begin{split}
			\underset{\alpha}{min}\quad &(\mathcal{L}_{val}^{std}(\omega^{*}(\alpha), \alpha),\ \lambda \mathcal{L}_{val}^{adv}(\omega^{*}(\alpha), \alpha)) \\
			s.t. \quad &\omega^{*}(\alpha)=argmin_{\omega}\ \mathcal{L}_{train}^{adv}(\omega,\alpha)
		\end{split}
	\end{cases}
\end{equation}

Specifically,~\cref{eq6} represents a bi-level optimization problem, where the lower-level formulation means updating network weights $\omega$ by minimizing adversarial loss $\mathcal{L}^{adv}_{train}(.)$ on the training set, and the upper-level formulation means updating architecture parameters to minimize both natural loss $\mathcal{L}^{std}_{val}(.)$ and adversarial loss $\mathcal{L}^{adv}_{val}(.)$ on the validation set. $\lambda$ represents a regularization coefficient. The adversarial loss can be optionally replaced by other robustness metrics.

Existing methods~\cite{dong2020adversarially,mok2021advrush,hosseini2021dsrna} transform the upper-level optimization to a single-objective optimization problem by summing the two objectives, with the fixed regularization coefficient, formulated as~\cref{eq111}.

\begin{equation}\label{eq111}
\underset{\alpha}{min}\quad \mathcal{L}_{val}^{std}(\omega^{*}(\alpha), \alpha)+\lambda \mathcal{L}_{val}^{adv}(\omega^{*}(\alpha), \alpha)
\end{equation}

\begin{table*}
	\centering
	\caption{Evaluation results of adversarially trained models on CIFAR-10 under white-box attacks. The best result in each column is in bold, and the second best is underlined. PGD$^{20}$ and PGD$^{100}$ refer to PGD attacks with 20 and 100 iterations, respectively. AA refers to the evaluation result after the standard group of AutoAttack method. All attacks are $l_{\infty}$-bounded with a total perturbation of 8/255.}
	\resizebox{\textwidth}{!}{
		\begin{tabular}{c|l|c|c|c|ccccc}
			\toprule
			Category&Model&Params&FLOPs&Natural Acc.&FGSM&PGD$^{20}$&PGD$^{100}$ & APGD$_{\rm CE}$ & AA\\
			\midrule
			\multirow{2}*{Hand-Crafted}&ResNet-18 & 11.2M &37.67M& 84.09\% & 54.64\% & 45.86\% & 45.53\% & 44.54\% & 43.22\%\\
			%\cmidrule(r){2-10}
			&DenseNet-121 & 7.0M&59.83M& 85.95\%& 58.46\%& 50.49\%& 49.92\% & 49.11\% & 47.46\%\\
			\midrule
			\multirow{2}*{Standard NAS}&{DARTS} & 3.3M&547.44M& 85.17\%& 58.74\%& 50.45\%& 49.28\% & 48.32\% & 46.79\%\\
			%\cmidrule(r){2-10}
			&PDARTS & 3.4M&550.75M& 85.37\%& 59.12\%& 51.32\%& 50.91\% & 49.96\% & 48.52\%\\
			\midrule
			\multirow{7}*{Robust NAS}&RobNet-free & 5.6M&800.40M& 85.00\%& 59.22\%& 52.09\%& 51.14\% & 50.41\% & 48.56\% \\
			%\cmidrule(r){2-10}
			&AdvRush & 4.2M&668.53M& 85.59\%&  59.98\%& 52.76\%&  52.55\% & 51.73\% & 49.28\%\\
			%\cmidrule(r){2-10}
			&{RACL} & 3.6M&568.86M& 83.97\%& 59.29\%& 52.13\%& 51.72\% & 51.24\% & 48.59\% \\
			%\cmidrule(r){2-10}
			&{DSRNA} & 2.0M&336.23M& 80.93\%& 54.49\%& 49.11\%& 48.89\% & 48.54\% & 44.87\% \\
			%\cmidrule(r){2-10}
			&{NADAR} & 4.4M&700.00M& \uline{86.23\%}& 60.46\%& \uline{53.43\%}& \uline{53.06\%} & \uline{52.64\%} & \uline{50.44\%} \\
			%\cmidrule(r){2-10}
			&{ABanditNAS-10} & 5.2M&794.11M& \textbf{90.64\%}& \textbf{81.31\%} & 50.51\%& 45.73\% & 29.31\% & 16.03\% \\
			%\cmidrule(r){2-10}
			& ARNAS(Ours)& 4.5M&1.27G& 85.92\%& \uline{62.45\%}& \textbf{55.87\%}& \textbf{55.43\%} & \textbf{54.84\%} & \textbf{52.66\%}\\
			\bottomrule
		\end{tabular}
	}
	\label{tab1}
\end{table*}

~\cref{eq111} can be optimized using gradient descent. However, as mentioned in~\cref{sec:intro}, scholars studying multi-objective optimization have shown that it would be better for identifying the Pareto front to always find a descent direction common to all criteria~\cite{desideri2012multiple,desideri2012mutiple}, while the fixed regularization coefficient cannot achieve this goal. To address this problem, we propose a multi-objective adversarial training method, based on MGDA~\cite{desideri2012multiple,sener2018multi}. MGDA is a gradient-based multi-objective optimization algorithm that either finds a common descent direction for all objectives and performs gradient descent, or does nothing when the current point is Pareto-stationary \cite{desideri2012multiple}. 

Specifically, we first need to determine the weights of all objectives dynamically. With the weights, all objectives can be optimized simultaneously. Thanks to the two objectives in our method, the process of dynamically determining the weights can be simplified as ~\cref{eq7},
\begin{equation}\label{eq7}
	\gamma^{*}=\underset{0\le\gamma\le1}{argmin}\left\|\gamma \theta+(1-\gamma)\bar{\theta}\right\|^{2}_{2}
\end{equation}
where $\theta=\nabla_{\alpha}\mathcal{L}_{val}^{std}(\omega^{*}(\alpha),\alpha)$, $\bar{\theta}=\nabla_{\alpha}\lambda\mathcal{L}_{val}^{adv}(\omega^{*}(\alpha), \alpha)$, $\nabla_{\alpha}$ denotes the gradient with respect to $\alpha$. Eq.~(\ref{eq7}) has an analytical solution, and can be calculated by~\cref{eq8}.
\begin{equation}\label{eq8}
	\gamma^{*}=max(min(\frac{(\bar{\theta}-\theta)^{T}\bar{\theta}}{\|\theta-\bar{\theta}\|^{2}_{2}},1),0)
\end{equation}
Using the obtained weights, the upper-level problem of~\cref{eq6} can be transformed to~\cref{eq9}, which ensures the simultaneous optimization of natural loss and adversarial loss, and can be implemented using gradient descent.
\begin{equation}\label{eq9}
	\underset{\alpha}{min}\quad \gamma^{*}\mathcal{L}_{val}^{std}(\omega^{*}(\alpha),\alpha)+(1-\gamma^{*})\lambda\mathcal{L}_{val}^{adv}(\omega^{*}(\alpha),\alpha)
\end{equation}

We implement the above algorithm using the second-order approximation~\cite{liu2018darts} of DARTS for competitive results. Meanwhile, the above algorithm involves the multiple computation of the same gradients of natural loss and adversarial loss, which can be saved and reused. In this way, the proposed algorithm does not require extra computational cost. At the end of search, an accurate and robust architecture can be obtained using the conventional process of DARTS.

\section{Experiments}\label{Experiments}
\subsection{Benchmark Datasets}\label{Benchmark Dataset}

Following the conventions of robust NAS community~\cite{guo2020meets,mok2021advrush}, CIFAR-10~\cite{krizhevsky2009learning}, CIFAR-100, SVHN~\cite{netzer2011reading} and Tiny-ImageNet-200~\cite{le2015tiny} are chosen as benchmark datasets. 

\subsection{Peer Competitors}\label{Peer Competitors}
State-of-the-art architectures from three categories are chosen as peer competitors. Specifically, the hand-crafted architectures are ResNet-18~\cite{he2016deep} and DenseNet-121~\cite{huang2017densely}. The architectures obtained by standard NAS are DARTS~\cite{liu2018darts} and PDARTS~\cite{chen2019progressive}. The architectures obtained by robust NAS are our main competitors, \textit{i.e.}, RobNet-free~\cite{guo2020meets}, AdvRush~\cite{mok2021advrush}, RACL~\cite{dong2020adversarially}, DSRNA~\cite{hosseini2021dsrna}, NADAR~\cite{li2021neural}, and ABanditNAS~\cite{chen2020anti}. 

\subsection{Parameter Settings}\label{Parameter Settings}
\textbf{Search Settings:} Following DARTS, we carry out architecture search on a small network consisting of 8 cells. The initial number of channels is set to 24, which is larger than conventions, aiming to keep the model capacity in the proposed search space similar to peer competitors for a fair comparison. Specifically, the numbers of channels divided by the two reduction cells are 24, 48, and 48, respectively. The total epoch is set to 50. To generate adversarial examples for the searching procedure, we use 7-step PGD with the step size of 2/255 and the total perturbation of 8/255. The regularization coefficient of the adversarial loss is set to 0.1. Remaining settings are the same as DARTS. We use momentum SGD to optimize the network weights $\omega$, with initial learning rate $\eta_{\omega}=0.025$ (annealed down to zero following a cosine schedule), 
momentum 0.9, and weight decay $3\times10^{-4}$. We use Adam as the optimizer for the architecture parameters, with initial learning rate $\eta_{\alpha}=3\times10^{-4}$, momentum $\beta=(0.5, 0.999)$ and weight decay~$10^{-3}$.

\textbf{Evaluation Settings:} To evaluate the searched architecture, we stack 20 cells to form a large network with initial number of channels of 64. Therefore, the number of channels divided by the two reduction cells are 64, 128, and 128, respectively. Following conventions~\cite{mok2021advrush}, we perform adversarial training using 7-step PGD with the step size of 0.01 and the total perturbation of 8/255 for 200 epochs. We use SGD to optimize the network, with the momentum of 0.9, and the weight decay of $1\times10^{-4}$. When evaluating on CIFAR-10 and CIFAR-100, the initial learning rate is set to 0.1. When evaluating on SVHN, the initial learning rate is set to 0.01. The learning rate is decayed by the factor of 0.1 at the 100-th and 150-th epoch. The batch size is set to 32. All the experiments are performed on a single NVIDIA GeForce RTX 2080 Ti GPU card.
\subsection{Results}\label{Results}

\subsubsection{White-box Attacks}\label{White-box Attacks}
We adversarially train the searched architectures and evaluate them under FGSM, PGD$^{20}$, PGD$^{100}$, and the standard group of AutoAttack (APGD$_{\rm CE}$, APGD$^{\rm T}$, FAB$^{\rm T}$, and Square)~\cite{croce2020reliable}. The results are shown in~\cref{tab1}. The results show that the ARNAS architecture achieves the highest adversarial accuracy among all competitors under PGD$^{20}$, PGD$^{100}$, and the standard group of AutoAttack, indicating that the ARNAS architecture is highly robust. Meanwhile, the natural accuracy of ARNAS outperforms all the competitors except DenseNet-121, ABanditNAS-10, and NADAR. Compared with them, the improvements of adversarial accuracy greatly exceeds the decrease of natural accuracy. Please note that AbanditNAS-10 only performs well under simple attacks such as FGSM. When the attacks get stronger, AbanditNAS-10 shows obviously lower accuracy than all other competitors, which indicates that AbanditNAS-10 is not actually trained to be robust. 
\begin{table}
	\centering
	\caption{Evaluation results of adversarially trained models on CIFAR-10 under transfer-based black-box attacks. In each row, the highest prediction accuracy except WRN-R (trained with additional 500k data) is in bold. In each column, the highest attack success rate (100\% - prediction accuracy) is underlined.}
	\resizebox{\columnwidth}{!}{
		\begin{tabular}{l|ccc|c}
			\toprule
			\diagbox[width=80pt]{Source}{Target}&\makecell[c]{WRN-R\\(500k data)}&ABanditNAS-10&AdvRush&ARNAS\\
			\midrule
			WRN-R (500k data)& - & 69.59\% &  68.99\% & \textbf{70.03\%}\\
			%\midrule
			ABanditNAS-10&84.82\% & -& 77.58\% &\textbf{78.39\%}\\
			%\midrule
			AdvRush&77.62\% & 68.83\%& -& \textbf{66.81\%}\\
			\midrule
			ARNAS&\underline{77.09\%} & \underline{67.80\%}& \underline{64.65\%}& -\\
			\bottomrule
		\end{tabular}
	}
	\label{tab2}
\end{table}

In addition, the ARNAS architecture has 1.27G FLOPs, which is significantly more than other architectures, even though their numbers of parameters are similar. Given that the ARNAS architecture is more robust, we infer that the number of parameters and the FLOPs are both the factors that affect the robustness. The conclusion also explains why previous studies get totally contradictory conclusions about the effect of model parameters on the robustness, i.e., some studies showed that more parameters can improve adversarial robustness~\cite{xie2019intriguing} while some others showed that more parameters may be harmful to adversarial robustness~\cite{huang2021exploring}. This may be because they ignore the influence of the FLOPs. Moreover, the large FLOPs are essentially caused by the special proportional relationship of channels designed in the proposed search space, which further demonstrates the effectiveness of the proposed search space. Specifically, the parameter size is proportional to the sum of channel numbers, while the FLOPs are proportional to the product of the channel numbers. The proposed search space keeps the sum of channel numbers similar to conventional search space, but the product of channel numbers is significantly larger, resulting in the architecture with similar parameter size but larger FLOPs.
\begin{table}
	\centering
	\caption{Evaluation results of adversarially trained models on CIFAR-100, SVHN, and Tiny-ImageNet under white-box attacks.}
	\resizebox{\columnwidth}{!}{
		\begin{tabular}{c|l|lll}
			\toprule
			Dataset&Model\qquad\quad\quad&Natural Acc.&FGSM\qquad\qquad&PGD$^{20}$\\
			\midrule
			\multirow{3}*{\quad CIFAR-100\quad}& ResNet-18& 55.57\%& 26.03\%& 21.44\%\\
			%\cmidrule(r){2-5}
			&PDARTS & \textbf{58.41\%}& 30.35\%& 25.83\%\\
			%\cmidrule(r){2-5}
			&ARNAS & 58.18\%& \textbf{32.60\%}& \textbf{29.54\%}\\
			\midrule
			\multirow{3}*{SVHN}&ResNet-18 & 92.06\%&  88.73\%& 69.51\%\\
			%\cmidrule(r){2-5}
			&PDARTS & 95.10\%& 93.01\%& 89.58\%\\
			%\cmidrule(r){2-5}
			&ARNAS & \textbf{95.84\%}& \textbf{94.43\%}& \textbf{92.02\%}\\
			\midrule
			\multirow{2}*{Tiny-ImageNet}&WideResNet & 52.10\%&  27.82\%& \textbf{24.83\%}\\
			%\cmidrule(r){2-5}
			\multirow{2}*{-200}&PDARTS & 45.94\%& 24.36\%& 22.74\%\\
			%\cmidrule(r){2-5}
			&ARNAS & \textbf{54.18\%}& \textbf{50.00\%}& 21.49\%\\
			\bottomrule
		\end{tabular}
	}
	\label{tab3}
\end{table}
\subsubsection{Black-box Attacks}\label{Black-box Attacks}
We conduct transfer-based black-box attacks, attacking the target model using adversarial examples generated by the source model. Adversarial examples from the source model are generated by PGD$^{20}$ with the total perturbation scale of 8/255. Except for aforementioned competitors, we perform extended experiments on WRN-34-R, which is the most robust variant of WideResNet found by~\cite{huang2021exploring}. WRN-34-R is trained using additional 500k data, so it shows the highest accuracy and robustness. We are interested in how ARNAS behaves when facing such a highly robust model. The results are presented in~\cref{tab2}.

The results show that ARNAS is more resilient against transfer-based black-box attacks than AdvRush and ABanditNAS. For example, when considering the model pair ABanditNAS-10 $\leftrightarrow$ ARNAS, ABanditNAS-10 $\rightarrow$ ARNAS achieves the attack success rate (100\% - prediction accuracy) of 21.61\%, while ARNAS $\rightarrow$ ABanditNAS-10 achieves the attack success rate of 32.20\%, where there is a gap of 10.59\%. Besides, when used as the target model (in each row), except for WRN-R that is trained with additional 500k data, ARNAS always has the highest prediction accuracy. When used as the source model (in each column), ARNAS always has the highest attack success rate, even higher than WRN-R. In conclusion, the black-box evaluation results further demonstrate the high robustness of the ARNAS architecture. Meanwhile, it proves that the ARNAS architecture does not unfairly benefit from the obfuscated gradients~\cite{athalye2018obfuscated} because the transfer-based black-box attacks do not use the gradients of target models.
\begin{figure*}
	\centering
	\includegraphics[width=\textwidth]{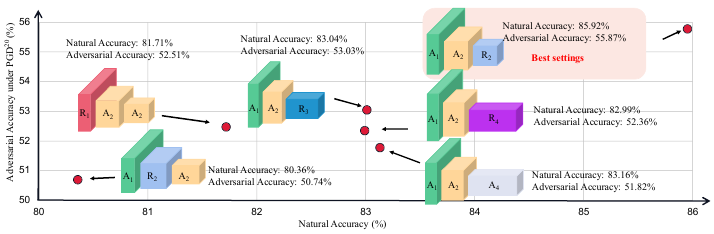}
	\caption{Visualization analysis of architectures in~\cref{tab4}. Character R refers Robust Cell, character A refers to Accurate Cell, and the subscript of A and R refers to filter settings. For example, R$_{4}$ refers to Robust Cell with the number of filters four times the initial.}\label{search_space_compare}
\end{figure*}
\subsubsection{Transferability to Other Datasets}\label{Transferability}
We transfer the ARNAS architecture to CIFAR-100, SVHN, and Tiny-ImageNet-200 to show its transferability. The results are shown in~\cref{tab3}. When transferred to CIFAR-100, the ARNAS architecture is far better than ResNet-18. Compared with PDARTS, the ARNAS architecture reaches competitive natural accuracy, while its FGSM accuracy and PGD$^{20}$ accuracy are significantly higher. When transferred to SVHN, the ARNAS perfroms best under all evaluation metrics. When transferred to Tiny-ImageNet-200, we replace ResNet-18 with a competitive model WideResNet-34-12~\cite{zagoruyko2016wide}. To our surprise, the ARNAS architecture reaches an unprecedented height in terms of natural and FGSM accuracy. Notably, the ARNAS architecture is almost never defeated under FGSM attack. Its FGSM accuracy is 22.18\% higher than WideResNet-34-12, and even close to the natural accuracy of WideResNet-34-12. The experimental results break the traditional prejudice that NAS-based architectures have weaker robustness than hand-crafted architectures as the dataset size or the task complexity increases~\cite{devaguptapu2021adversarial}.

\subsubsection{Ablation Study}\label{Ablation Study}
\textbf{Ablation study of search Space.} We try all possible placement of the Accurate Cells and the Robust Cells while the placement of the Reduction Cells is the same as the NASNet search space. When the placement is determined, we further study on the different settings of the number of filters. The results are shown in~\cref{tab4}. To represent the placement of the cells, we use A for Accurate Cell and R for Robust Cell. For example, A-A-R means we place the Accurate Cells before the second Reduction Cell and the Robust Cells after the second Reduction Cell, which is the same as the proposed search space described in~\cref{Search Space}. For filter settings, we use N$_{1}$-N$_{2}$-N$_{3}$ to represent the proportional relationship of the number of filters. For example, 1-2-4 means the number of filters between the first Reduction Cell and the second Reduction Cell is two times the initial number of filters, and the number of filters after the second Reduction Cell is four times the initial number of filters, which is adopted by the NASNet search space.

\begin{table}
	\captionsetup{width=0.45\textwidth}
	\centering
	\caption{Ablation study of search space on CIFAR-10.}
	\resizebox{\columnwidth}{!}{
		\begin{tabular}{c|cc|cr}
			\toprule
			Row Number&Placement&Filter Setting&Natural Acc.&PGD$^{20}$\\
			\midrule
			1&A-A-R & 1-2-2 & \textbf{85.92\%} & \textbf{55.87\%}  \\
			%\midrule
			2&R-A-A & 1-2-2 & 81.71\% & 52.51\% \\
			%\midrule
			3&A-R-A & 1-2-2 & 80.36\%& 50.74\%\\
			%\midrule
			4&A-A-R & 1-2-3& 83.04\%& 53.03\% \\
			%\midrule
			5&A-A-R & 1-2-4& 82.99\%& 52.36\%\\
			%\midrule
			6&A-A-A & 1-2-4& 83.16\%& 51.82\%\\
			\bottomrule
		\end{tabular}
	}
	\label{tab4}
\end{table}

\begin{table}
	\centering
	\caption{Ablation study of search strategy on CIFAR-10.}
	\resizebox{\columnwidth}{!}{
		\begin{tabular}{c|cc|cc}
			\toprule
			Row Number&ARNAS Search Space&Multi-Objective&Natural Acc.& PGD$^{20}$\\	
			\midrule
			1&\CheckmarkBold&\XSolidBrush&85.04\% & 53.72\% \\
			%\midrule
			2& \CheckmarkBold& \CheckmarkBold&85.92\% & 55.87\% \\
			\bottomrule
		\end{tabular}
	}
	\label{tab5}
\end{table}
When the filter setting is fixed to be 1-2-2 (experiments 1, 2 and 3), the best result is achieved when we set the placement to be A-A-R (experiment 1). So we fix the placement to be A-A-R and conduct further experiments. When the placement is fixed to be A-A-R (experiments 1, 4 and 5), the best result is achieved exactly when we set the number of filters to be 1-2-2 (experiment 1), which is adopted in the previous experiments. Therefore, we construct the proposed search space with the placement of A-A-R and the filter setting of 1-2-2. Compared with conventional search space (placement of A-A-A and filter setting of 1-2-4), the neural architecture searched from the proposed search space achieves 2.23\% higher natural accuracy and 2.78\% higher PGD$^{20}$ accuracy. We also provide a visualization analysis in~\cref{search_space_compare} for the above experimental results. As shown in the figure, a well-designed setting can largely improve the accuracy and the robustness (top right corner of the figure). 

\textbf{Ablation study of search strategy.} The innovation of the proposed search strategy lies in the proposed multi-objective adversarial training method. We compared it with the previous method (i.e., sum of two objectives with a fixed regularization term). The results are shown in~\cref{tab5}. Using the proposed multi-objective adversarial training method, the searched architecture get both higher accuracy and robustness. Commonly, the optimization of architecture based on gradient descent does not guarantee convergence to the optimal solution. It is possible to see that the two conflicting objectives get both higher. The results demonstrate that the proposed multi-objective adversarial training method is more effective than the previous one.

\subsubsection{Analysis on Architectural Ingredients of Accurate and Robust Neural Networks}\label{Architectural Ingredients}

\begin{figure}
	\includegraphics[width=\columnwidth]{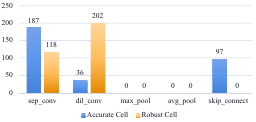}
	\caption{Statistical analysis on 40 neural architectures searched by the proposed method. The numbers of times of the operations selected by Accurate Cell and Robust Cell are recorded.}\label{figure_statistic}
\end{figure}

To get more insights of accurate and robust architectures, we repeat the proposed method for four times with different random seeds. For every execution, the top-10 architectures are recorded. Then a statistical analysis of these 40 neural architectures is carried out, and the results are shown in~\cref{figure_statistic}. We find that the ARNAS architectures tend to deploy very different structures for Accurate Cells and Robust Cells. Specifically, the Accurate Cells prefer separable convolutions while employing a few dilated convolutions and skip connections. However, the Robust Cells prefer dilated convolutions while employing a few separable convolutions and no skip connection. Such kind of neural architectures are impossible to be found in the conventional search space, because the conventional search space limits the cells near the input and output to be the same~\cite{zoph2018learning}.

Next, we will analyze why Accurate Cells and Robust Cells contribute to accuracy and robustness, respectively. Please note that the skip connections are not in our consideration, because they have effects on the training process by accelerating gradient propagation~\cite{glorot2010understanding}, instead of improving the learning ability of the architecture. On the one hand, it is a consensus that more learnable parameters are beneficial for accuracy~\cite{yarotsky2017error}. Existing practice is also consistent with the consensus. For example, architectures searched by popular standard NAS algorithms, such as DARTS and PDARTS, are mainly composed of separable convolutions, which requires the maximum number of parameters among all optional operations. The cell structures near the input in our experiments are in line with the consensus and the previous practice. This is exactly why we name these cells as Accurate Cells. On the other hand, it is also empirically and theoretically proved that more learnable parameters are harmful to the robustness~\cite{huang2021exploring}. Among the optional operations, the dilated convolutions fix some weights to be zero, so they have fewer learnable parameters. As a result, perturbations of input are difficult to change the output, resulting in the stronger robustness. In our experiments, the cells near the output tend to employ dilated convolutions instead of separable convolutions, and the robustness may be enhanced. This is exactly why we name these cells as Robust Cells. In conclusion, although accuracy and robustness are conflicting objectives, the cell structures in different positions play different roles for accuracy and robustness, and the architectures can obtain both accuracy and robustness by deploying very different structures in different positions.

The conclusion has great guiding significance on both hand-crafting and automatically designing of accurate and robust architectures. To our knowledge, though it is common to design architectures by stacking repeated structures nowadays, few people try to stack very different structures in different positions, which may greatly limit the performance of the designed architectures.
\section{Conclusion}\label{Conclusion}
In this work, we propose the ARNAS method to search for accurate and robust neural architectures after adversarial training automatically. Specifically, we first design the ARNAS search space specially for adversarial training through experiments, which empirically improve the accuracy and the robustness of the searched neural architectures. Then we design a differentiable multi-objective search strategy, searching for accurate and robust architectures by performing gradient descent towards a common descent direction of natural loss and adversarial loss. We evaluate the searched architecture under various adversarial attacks on various benchmark datasets. The strongest robustness and outstanding accuracy of the searched architecture demonstrate the superiority of the proposed method. Meanwhile, the unexpected transferability break the traditional prejudice that NAS-based architectures are inferior to hand-crafted architectures as the task complexity increases in robustness scenario.  Besides, we reach an important conclusion that the cell structures in different positions of the architectures play different roles for accuracy and robustness, and the architectures can obtain high accuracy and high robustness simultaneously by deploying very different cell structures in different positions. The conclusion has great guiding significance on both hand-crafting and automatically designing of accurate and robust neural architectures. 

\section*{Acknowledgments}
This work was supported by National Natural Science Foundation of China under Grant 62276175.

{
    \small
    \bibliographystyle{ieeenat_fullname}
    \bibliography{main}
}


\end{document}